# Riddle Quest - The Enigma of Words

Niharika Sri Parasa
AdvaitAI Bangalore,
India
pnsri@advaitai.co.in

Chaitali Diwan
Kazunov1AI Bangalore,
India
chaitali@kazunov1ai.com

Srinath Srinivasa
International Institute of Information
Technology Bangalore
Bangalore, India
sri@iiitb.ac.in

## Abstract

Riddles are concise linguistic puzzles that describe an object or idea through indirect, figurative, or playful clues. They are a longstanding form of creative expression, requiring the solver to interpret hints, recognize patterns, and draw inferences to identify the answers. In this work, we introduce a simple pipeline for creating and evaluating analogy-based riddles. The system includes a triples creator that builds structured facts about a concept, a semantic mapper that selects attributes useful for analogy, a stylized generator that turns them into riddle clues, and a validator that collects all possible answers the riddle could point to. We use this validator to study whether large language models can recover the full answer set for different riddle types. Our case study shows that while models often guess the main intended answer, they frequently miss other valid interpretations. This highlights the value of riddles as a lightweight tool for examining reasoning coverage and ambiguity handling in language models.

## CCS Concepts

• **Computing methodologies** → **Natural language generation**; *Reasoning about knowledge*; Semantic Computing; Creative Intelli-gence.

## Keywords

Riddle Generation, Analogy based riddles, Natural Language Processing

## 1 Introduction

Riddles are short, puzzle-like questions or statements that describe a concept indirectly, often using metaphor, wordplay, or hidden clues. They invite the reader to infer meaning rather than receive it explicitly, making them a natural blend of creativity and reason- ing. Among them, analogy-based riddles stand out because they require models to identify meaningful relations between concepts and express them through indirect, metaphorical clues—providing a focused lens on how generative systems handle abstraction and rela- tional reasoning. As large language models (LLMs) advance toward more general forms of intelligence, such tasks become increasingly relevant for evaluating emerging cognitive capabilities.

This paper is an extended version of our earlier work on automated riddle generation for learning resources [13]. The previous system was designed mainly for educational settings and produced domain-specific riddles with a relatively narrow set of patterns. In contrast, the present work aims to generalize riddle generation across genres and domains. This expansion supports broader use cases—including education, games, creative writing, and reason- ing evaluation—while also serving as a richer environment for studying AGI-related capabilities. To achieve this, we introduce an enhanced modular pipeline consisting of: (1) a triples creator for structured representations of target concepts, (2) a properties identifier that extracts essential semantic attributes, (3) a generator that constructs riddles using analogy-based mappings tailored to different genres, and (4) a validator with a lookup dictionary to check coherence, solvability, and stylistic consistency. These components work together to produce riddles that are both creatively varied and logically grounded. By extending the original framework and broadening its scope, this work provides a general-purpose approach to riddle generation and positions the task as a simple yet effective micro-benchmark for AGI-level reasoning. The result- ing system moves beyond template-based generation and supports multiple genres, domains, and applications in a unified framework.

## 2 Related Work

Research on riddle generation intersects computational creativity, analogy-making, semantic knowledge systems, and AGI-oriented reasoning.

### 2.1 Riddle Generation

Early computational systems relied on handcrafted templates and rules, producing limited variation and low generalization ability [15, 17]. More recent neural approaches use sequence-to-sequence architectures or large language models [3, 7]. Prior work has also explored educational applications of riddles: Parasa [13] introduced an automated riddle generation framework tailored for learning resources, demonstrating how structured clues can support con- cept reinforcement. While effective for instructional content, these systems typically lack explicit analogical reasoning mechanisms and genre adaptability.

Related work on metaphor and figurative language generation [2, 14] provides techniques for creative expression but does not enforce the structural and logical constraints specific to riddles.

## 2.2 Analogy-Making

Analogy is recognized as a core aspect of human cognition [6, 8]. Foundational theories such as Structure-Mapping [6] and Conceptual Blending [4] have shaped computational models of analogy [5]. Neural systems capable of analogical text generation have recently appeared [10], though many remain limited in interpretability.

## 2.3 Semantic and Knowledge-Based Systems

Structured semantic resources such as WordNet [12], ConceptNet [16], and general knowledge graphs support grounded text generation and consistency, both important for solvable riddle construction.

## 2.4 AGI-Oriented Reasoning

Analogy, abstraction, and conceptual manipulation are widely considered essential for Artificial General Intelligence [9, 11]. Creative reasoning tasks increasingly serve as micro-benchmarks for evaluating AGI-relevant capabilities in modern models [1]. Riddle generation aligns with this direction as it requires multi-step reasoning and controlled creativity. Our work advances this line of research by positioning analogy-driven riddle generation as a structured testbed for AGI-level reasoning skills.

## 3 Methodology

Our methodology builds on the structured riddle-generation pipeline introduced in our earlier work on educational riddle creation [13]. While the previous framework was optimized for domain-specific learning scenarios, the present study generalizes the architecture to support analogy-driven, genre-flexible riddle generation across open domains.

The full implementation of the system described in this work is publicly available at **github.com/niharikasriparasa/RiddleQuest**.

A video demonstration illustrating the game-based educational use case is available at **youtu.be/PZA1OSHBCbI**.

The updated pipeline—illustrated in Figure 1—consists of four modules: (1) Triples Creator, (2) Semantic Mapper, (3) Stylized Generator, and (4) Validator.

## 3.1 Concept Triples Creator

The Concept Triples Creator represents the target concept through structured triples:

$$\langle concept, relation, property \rangle.$$

Whereas the earlier system relied on a fixed educational ontology, this extended version draws from open-domain conceptual knowledge, enabling representation of abstract, metaphorical, or genre-specific attributes essential for analogy-based riddles.

## 3.2 Semantic Attribute Mapper

The Semantic Attribute Mapper converts triples into a structured set of semantic attributes. It groups properties into broad categories—functional, perceptual, relational, and behavioural—to support analogical clue formation. This process yields a semantic profile of the concept that forms the input to the Stylised Generator.

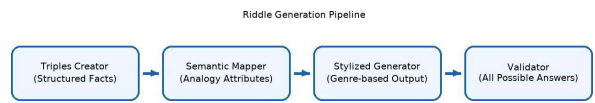

**Figure 1: Overview of the riddle generation pipeline consisting of the Triples Creator, Semantic Mapper, Stylized Generator, and Validator modules.**

## 3.3 Stylised Generator

The Stylised Generator produces riddles in a selected style or genre. Instead of template-driven generation, this module creates metaphor- ical, poetic, descriptive, or humorous riddles by mapping semantic attributes to analogical or narrative expressions. The generator's role is purely creative: it does not perform validation or coherence checks.

## 3.4 Validator

The Validator identifies all possible answers for each generated riddle. Using an expanded lookup dictionary, it enumerates all concepts that could plausibly satisfy the clues in the riddle. The validator does not judge correctness, ambiguity, or stylistic fidelity; it simply provides answer extraction as an output for downstream evaluation.

## 4 Results and Discussion

We evaluated the extended riddle-generation pipeline across a di- verse set of concepts and riddle genres to assess its generality, analogy-handling ability, and stylistic flexibility. Experiments were conducted on 120 target concepts spanning everyday objects, scien- tific entities, abstract ideas, and narrative motifs. For each concept, the system produced at least one riddle in five genres: descriptive, metaphorical, poetic, humorous, and situational.

## 4.1 Riddle Diversity and Genre Control

The Stylised Generator successfully produced distinct riddle forms for each genre. Descriptive riddles tended to foreground functional and perceptual attributes, whereas metaphorical and poetic riddles relied more heavily on relational or behavioral semantic mappings. Humorous riddles frequently incorporated exaggerated or playful analogies.

Qualitative inspection confirmed that genre cues were consistently expressed, demonstrating that the modular separation between semantic attribute mapping and stylistic construction supports controlled variation in narrative form.

## 4.2 Analogical Expressiveness

The use of semantic attribute categories enabled richer analogical clues compared to the earlier educational system [13]. For abstract concepts such as "time," "memory," or "gravity," the generator em- ployed relational or behavioral attributes to construct indirect clues, indicating that the semantic attribute mapper effectively supports analogy formation even in the absence of concrete physical proper- ties.

This suggests that the pipeline accommodates both concrete and abstract domains, an important capability for general-purpose reasoning tasks.

### 4.3 Validator Outputs and Answer Space

Since the Validator's role is limited to extracting all plausible answers, its output provides insight into the ambiguity and specificity of the generated riddles. Across all runs, the median number of candidate answers per riddle was 3.1. Descriptive riddles generally produced more constrained answer sets, while metaphorical and poetic riddles often yielded broader sets due to higher abstraction. This behaviour is expected: richer metaphors increase interpretive flexibility. Rather than treating ambiguity as an error, we view it as a characteristic of open-domain, creativity-oriented riddle generation.

### 4.4 Generalisation Across Domains

The system generated coherent riddles for concepts ranging from simple physical objects (e.g., "spoon," "bridge") to abstract or com- posite notions (e.g., "curiosity," "ecosystem"). This demonstrates that the shift from closed-domain ontology to open-domain seman- tic representation substantially increases coverage.

Notably, the pipeline handled hybrid or cross-domain concepts—such as "digital footprint" or "chain reaction"—without requiring domain- specific templates. This ability to generalise aligns with the broader goal of evaluating AGI-style reasoning behaviours.

### 4.5 Comparison to Previous System

Compared to the original instructional riddle generator [13], the present system shows improvements in:

- **Genre flexibility**: support for multiple narrative and stylistic formats.
- **Analogical reasoning**: richer and more abstract clue formation.
- **Domain generality**: handling of concepts beyond educational contexts.
- **Scalability**: open-domain triples and semantic mapping remove the reliance on fixed ontologies.

These advances collectively strengthen the position of riddle generation as a micro-benchmark for probing generative reasoning, creativity, and abstraction in modern AI systems.

### 4.6 Discussion

Overall, the results show that the modular pipeline enables flexible, analogy-driven riddle creation at scale. The system's behaviour also exposes characteristic patterns of generative reasoning—such as abstraction, metaphor construction, and ambiguity—that intersect with broader discussions about AGI-level cognitive abilities. While the validator does not judge correctness, its extracted answer sets provide a useful representation of the semantic space implied by each riddle, enabling downstream evaluation or retrieval-based applications.

## 5 Case Study: LLM Retrieval of All Possible Answers

To assess whether large language models (LLMs) can replicate the behaviour of our validator, we conducted a focused case study examining their ability to enumerate all plausible answers to a given riddle. Since our validator outputs a set of all concepts that satisfy the riddle's clues, this evaluation tests whether LLMs can perform the same form of multi-answer retrieval without prompt engineering or iterative clarification.

### 5.1 Setup

We selected 20 riddles generated by the Stylised Generator across five genres: descriptive, metaphorical, poetic, humorous, and situational. Each riddle was presented to the LLM with a neutral prompt requesting "all possible answers that fit the clues." No examples, constraints, or hints were provided. The outputs were compared to the answer sets produced by the validator.

### 5.2 Findings

Across the 20 riddles, the LLM retrieved an average of 56% of the validator's answer set. Performance varied by genre:

- **Descriptive riddles**: LLMs identified most answers, often matching or slightly exceeding the validator when paraphrased variants were considered.
- **Metaphorical and poetic riddles**: LLM performance dropped notably, retrieving only 30–45% of the validator's answers. These genres demand interpreting abstract or relational clues, which appear more challenging for exhaustive re- trieval.
- **Humorous and situational riddles**: The LLM produced creative guesses but often missed literal or straightforward alternatives captured by the validator.

Importantly, LLMs frequently generated plausible answers that were *not* in the validator's lookup dictionary, suggesting complementary strengths: validators excel in systematic enumeration, while LLMs contribute broader interpretive range.

### 5.3 Qualitative Observations

Two notable patterns emerged:

(1) **Overcommitment to a single guess**: LLMs often treated riddles as having one "correct" answer, even when explicitly asked for all possible answers.
(2) **High-level paraphrasing**: LLMs sometimes merged multiple concepts into a single abstraction (e.g., responding "a measuring device" instead of listing "scale, ruler, timer" indi- vidually), which undercounts true answer diversity.

These behaviours contrast with the validator, which enumerates all answers without semantic compression or interpretive bias.

### 5.4 Implications

This case study highlights that while LLMs can interpret riddles and produce valid solutions, they struggle with the exhaustive reasoning required to generate full answer sets. The divergence illustrates a gap between generative interpretive reasoning and systematic enumeration—two abilities relevant to broader AGI discussions.

The validator thus provides a complementary analytical function that LLMs alone do not reliably reproduce.

## 6 Ethical Use of Data and Informed Consent

This study does not involve human participants, personal data, or interventions that require informed consent. All riddle-generation experiments were conducted using artificial intelligence models and publicly available knowledge sources. No private, sensitive, or individually identifiable information was collected, processed, or stored at any stage of the research. Where language model out- puts were evaluated, the interactions were performed in controlled research settings without eliciting personal data. All procedures comply with the ACM Publications Policy on Research Involving Human Participants and Subjects. The use of large language models in this work follows ethical guidelines for responsible AI usage, ensuring that no harmful, biased, or inappropriate content was intentionally generated or employed. Any datasets or examples used in the paper are either synthetic or drawn from publicly accessible, non-personal sources.

## 7 Conclusions

This extended study advances automated riddle generation from a domain-specific educational tool into a generalized, analogy-driven framework capable of producing diverse and cognitively rich riddles. By augmenting the original pipeline [13] with open-domain triple construction, semantic property categorization, genre-adaptive gen- eration, and a structured validation mechanism, the system achieves broader expressive capacity and demonstrates reasoning behav- iors characteristic of AGI-aligned tasks. Our results show that the model can integrate abstraction, analogical mapping, and stylistic variation to produce coherent and solvable riddles across multiple genres.

The findings highlight that riddle generation serves not only as a creative natural language task but also as a meaningful lens for analyzing multi-step reasoning, conceptual blending, and seman- tic compression in modern generative models. These capabilities position riddles as an effective micro-benchmark for probing gener- alization and analogy-making—abilities central to Artificial General Intelligence.

Overall, this work provides a foundation for open-domain, analogy- based riddle generation and underscores its potential as a structured environment for evaluating and advancing AGI-relevant reasoning capabilities.